\documentclass{esannV2}
\usepackage{graphicx}
\usepackage[latin1]{inputenc}
\usepackage{amssymb,amsmath,array}
\usepackage{subfigure}
\usepackage{sidecap}

\begin{document}

\title{Personalizing human activity recognition models using incremental learning}

\author{Pekka Siirtola, Heli Koskim\"{a}ki and Juha R\"{o}ning
%
%
\vspace{.3cm}\\
Biomimetics and Intelligent Systems Group,\\ P.O. BOX 4500, FI-90014, University of Oulu, Oulu, Finland\\
\{pekka.siirtola, heli.koskimaki, juha.roning\}@oulu.fi
}

\maketitle


%
%
%
\hyphenation{manu-scripts manu-script ext-re-mums user-in-de-pen-dent user-de-pen-dent in-de-pen-dent mo-del in-de-pen-dent}

\begin{abstract}
In this study, the aim is to personalize inertial sensor data-based human activity recognition models using incremental learning. 
At first, the recognition is based on user-independent model. However, when personal streaming data becomes available, the incremental learning-based recognition model can be updated, and therefore personalized, based on the data without user-interruption. 
The used incremental learning algorithm is Learn++ which is an ensemble method that can use any classifier as a base classifier. In fact, study compares three different base classifiers: linear discriminant analysis (LDA), quadratic discriminant analysis (QDA) and classification and regression tree (CART). Experiments are based on publicly open data set and they show that already a small personal training data set can improve the classification accuracy. Improvement using LDA as base classifier is 4.6 percentage units, using QDA 2.0 percentage units, and 2.3 percentage units using CART. However, if the user-independent model used in the first phase of the recognition process is not accurate enough, personalization cannot improve recognition accuracy. 
\end{abstract}


%
\section{Problem statement and related work}
\label{problem}

This study focuses on human activity recognition based on inertial sensor data collected using smartphone sensors. One of the main challenge of the field is that people are different: they are unique for instance in terms of physical characteristics, health state or gender. All of these have an effect to the data that are collected for model training. In fact, it is shown that user-independent models do not work accurately for instance if they are trained with healthy study subjects and tested with subjects who have difficulties to move \cite{albert2012using}. Thus, if the aim is to train a model that works with everybody, the focus of research should be on personal and personalized prediction models instead of user-independent models. However, the challenge of personal and personalized models is that they require personal training data. This normally would require an extensive, separate data collection session for each user making the approach unusable out-of-the-box.

There are some attempts to personalize the recognition process without user-interruption. Siirtola \textit{et. al} \cite{siirtola20ESANN} presented a two step approach especially designed for devices that include several different types of sensors. In the first step, user-independent model that uses data from all sensors was used to label personal streaming data. This information is then used to build a light, and energy efficient personal model that uses only a sub-set of available sensors. Using the approach, recognition accuracy can be improved but the problem of the approach are that personalization is based on model re-training. Therefore, all the streaming data needs to be stored to be able to use it for model training. This obviously is problematic as it requires a lot memory. There are also other studies, where recognition models are personalized in order to improve the model accuracy. In \cite{Fallahzadeh} model personalization was based on transfer learning algorithm. Also this article shows the importance of model personalization, it was shown that the recognition accuracies of personalized models are significantly better than the one's based on user-independent models. 

Incremental learning refers to recognition methods that can learn from online information and adapt to new environments. The advantage is that this adaptation can be done without model re-training and user-interruption. Instead, models can be updated automatically based on streaming data \cite{gepperth2016incremental}. There are some studies where incremental learning is used to recognize human activities based on wearable sensor data \cite{wang2012incremental,Mo2016,Ntalampiras}. These studies show that inertial sensor-based models benefit from incremental learning as it improves the recognition rates. However, in these studies 
the focus has not been on personalizing the recognition models which is the aim in our study. 
In fact, in our study does not only show that personalization based on incremental learning improves the recognition rates compared to results of user-independent model, it also compares three base classifiers:  LDA, QDA and CART. The experiments are made using publicly open data set.




\section{Experimental dataset}
\label{data}
The experiments of this study are based on the
publicly open data set presented in \cite{Shoaib} which contains data from seven physical activities (walking, sitting, standing, jogging, biking, walking upstairs and downstairs). The data were collected using a smartphone and from five body locations but in this study only the data collected from waist were used. The data ware collected from 3D accelerometer, 3D gyroscope, and 3D magnetometer using sampling rate of 50Hz. This study uses data from accelerometer and gyroscope. Data set contains measurements from 10 study subjects. However, apparently one of the study subjects had placed sensor in different orientation than others making the data totally different to other subjects' data. Thus, person's data were not used in the experiments.

Window size of 4.2 second with 1.4 second slide was used in the study. From these windows, features were extracted. This study uses features that are commonly used in activity recognition studies including standard deviation, minimum, maximum, median, and different percentiles (10, 25, 75, and 90). Moreover, the sum of values above or below percentile (10, 25, 75, and 90), square sum of values above of below percentile (10, 25, 75, and 90), and number of crossings above or below percentile (10, 25, 75, and 90) were extracted and used as features. In addition, features from frequency domain, for instance sums of small sequences of Fourier-transformed signals, were extracted. Features were extracted from raw accelerometer and gyroscope signals, magnitude signals and signals where two out of three accelerometer and gyroscope signals were combined. Altogether 244 features were extracted.

\section{Personalizing recognition without user-interruption}
\label{aim}


The aim of this study is to show that incremental learning can be used to personalize human activity recognition models, and therefore, to improve recognition accuracy without user-interruption.
In this study, incremental learning is based on Learn++ algorithm \cite{polikar2001learn++}. Learn++ is an ensemble method where the idea is to process incoming streaming data as chunks. For each chunk a new group of weak base models are trained and combined to a group of previously trained base models through weighted majority voting as ensemble model \cite{hammerchoosing}. 

The reason to use Learn++ is that in \cite{hammerchoosing} it is shown that it is not only accurate but also less complex than many other algorithms. Therefore, it is suitable to be implemented devices that do not have much memory and calculation capacity such as wearable sensors. Moreover, Learn++ can use any classifier as base classifier. Therefore, it was possible to select such base classifiers that are used also in the previous human activity recognition studies. In fact, this article compares three different base classifiers: CART, LDA and QDA.

\begin{SCfigure}
	\centering
		\includegraphics[width=8.6cm,keepaspectratio = true]{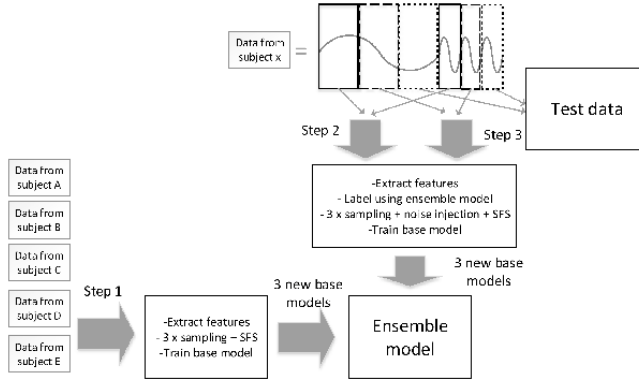}
\caption{Process to personalize human activity recognition based on Learn++ algorithm.}
\label{process}
\end{SCfigure}

A method to personalize recognition process is explained using an example in Figure \ref{process}. In the approach leave-one-out method is used: model training starts from user-independent data set including data from all study subjects except one. Data from one person in turn is used for personalization and testing. 
In the figure, the aim is to build personalized model for subject x and, in this example, data consists of measurements from 2 classes. User-independent model is therefore based on data from subjects A-E. Subject x's data are divided into three parts (solid, dashed and dotted lines) so that each part contains the same amount of data from each activity. Two parts (solid and dashed) are used for personalizing the recognition model and one part (dotted) is for testing.

Step 1 to train Learn++ based recognition model is to randomly sample data from subjects A-E to build a user-independent base model, best features for this are selected using SFS (sequential forward selection). New base model is then trained based on them and it is added to the ensemble of models. Ensemble model is then tested using test data set, which is subject x's data bordered with dotted line. Step 1 is repeated $n$ times, in this case $n=3$. 
Step 2 is to start personalizing the ensemble model by extracting features from the first part of subject x's data which is bordered with solid line. This data is labeled using ensemble model. Problem with the data chunks used to personalize the recognition process is that they are small. Thus, they do not contain much variation leading easily to over-fitted models. To avoid over-fitting, noise injection method presented in \cite{cidm2016} is applied to training data sets to increase the size of training data and increase it's variation. After this, a data set used in model training is selected based on random sampling and SFS is applied to it to select best features. New base model is then trained using the selected features and added to ensemble model, which is again tested using subject x's data bordered with dotted line. Step 2 is repeated three times. Step 3 is to do the same with subject x's data bordered with dashed line. Also step 3 is repeated three times. Therefore, in the end the Learn++ conducts of nine base models. 


\section{Experiments and discussion}
\label{experiments}

\begin{figure}[h]
\begin{center}
 		\subfigure[Subject 1.]{\label{f1}\includegraphics[width=3.75cm]{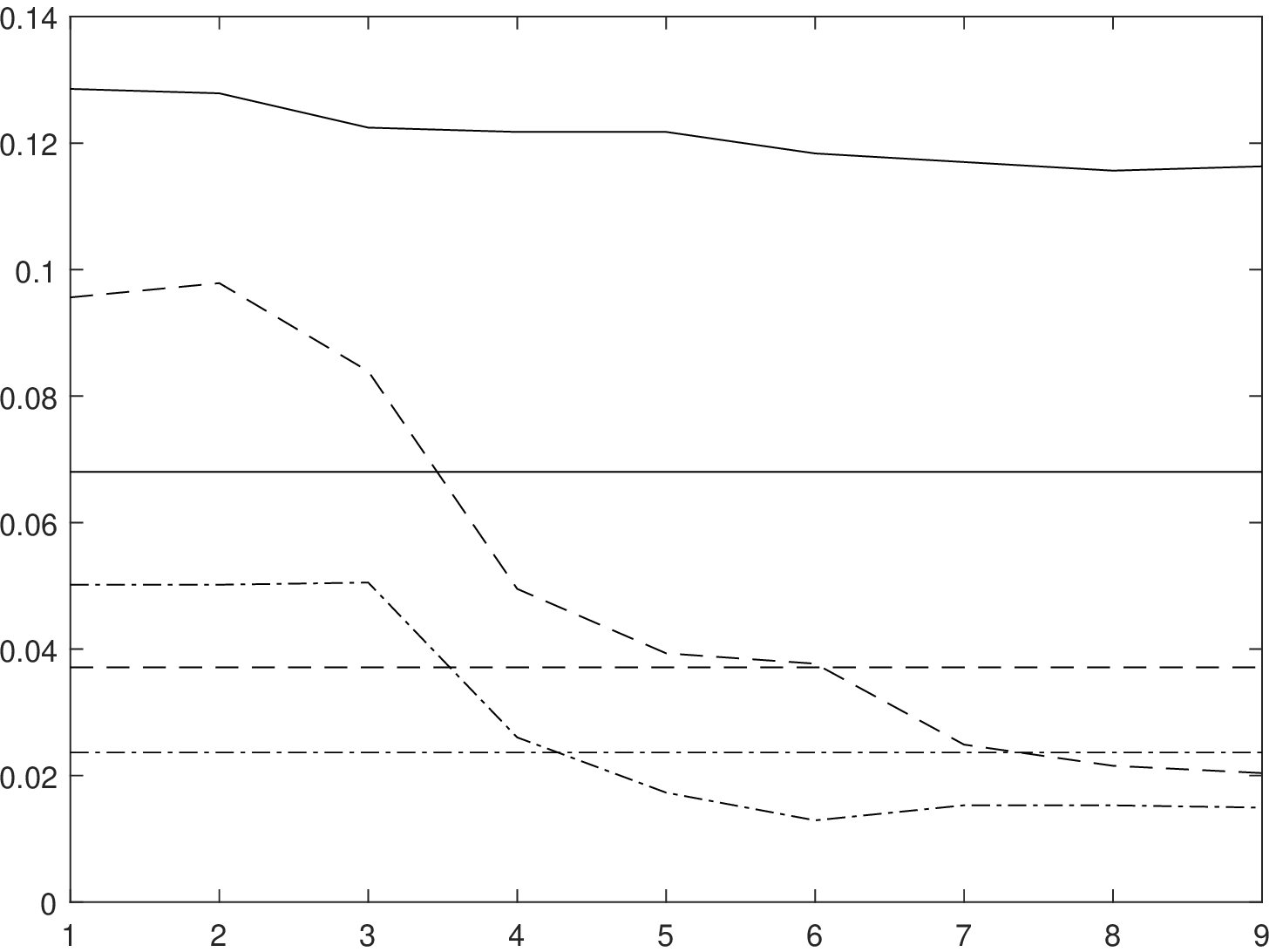}}
		\subfigure[Subject 2.]{\label{f2}\includegraphics[width=3.75cm]{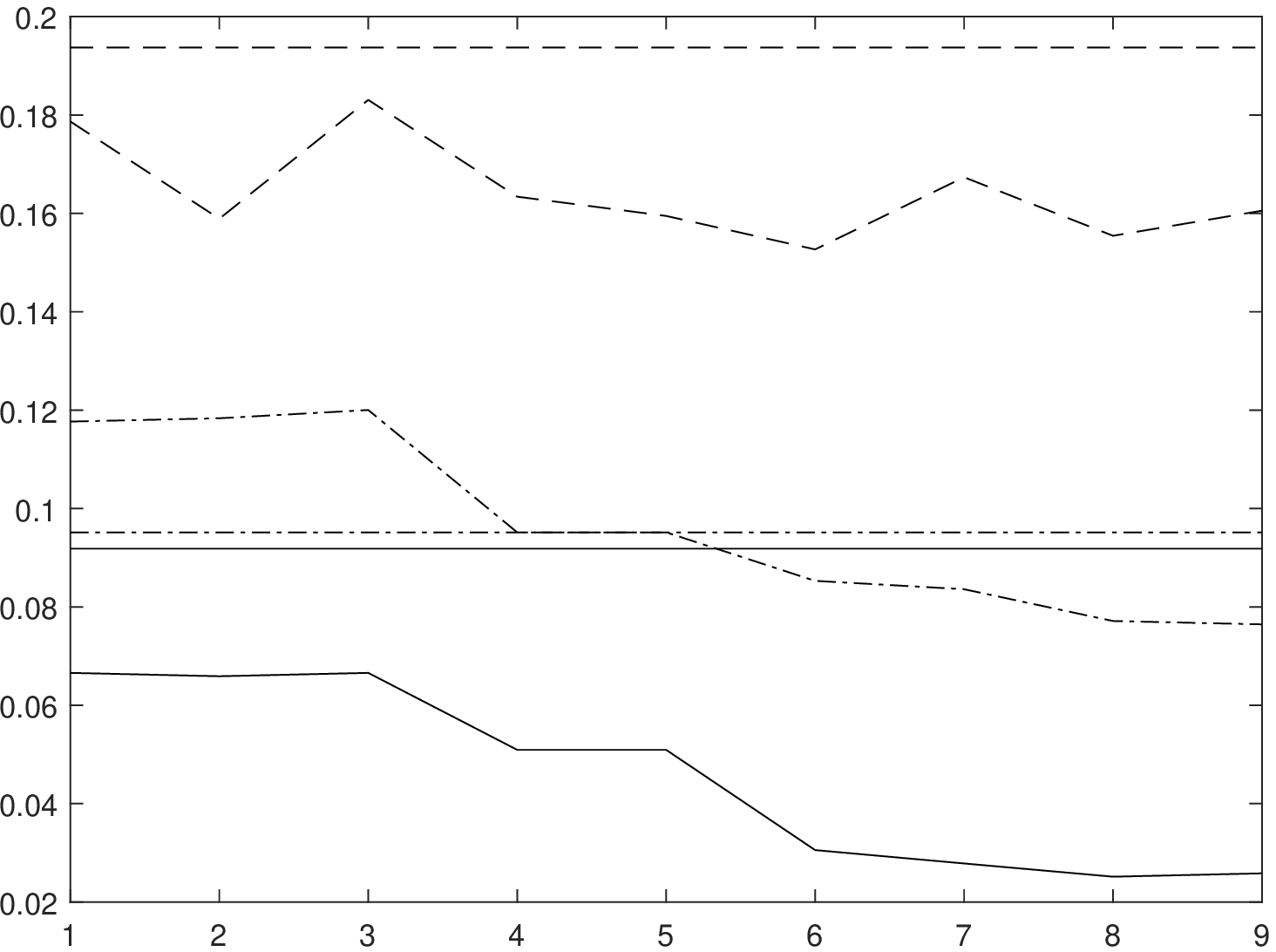}}
         \subfigure[Subject 3.]{\label{f3}\includegraphics[width=3.75cm]{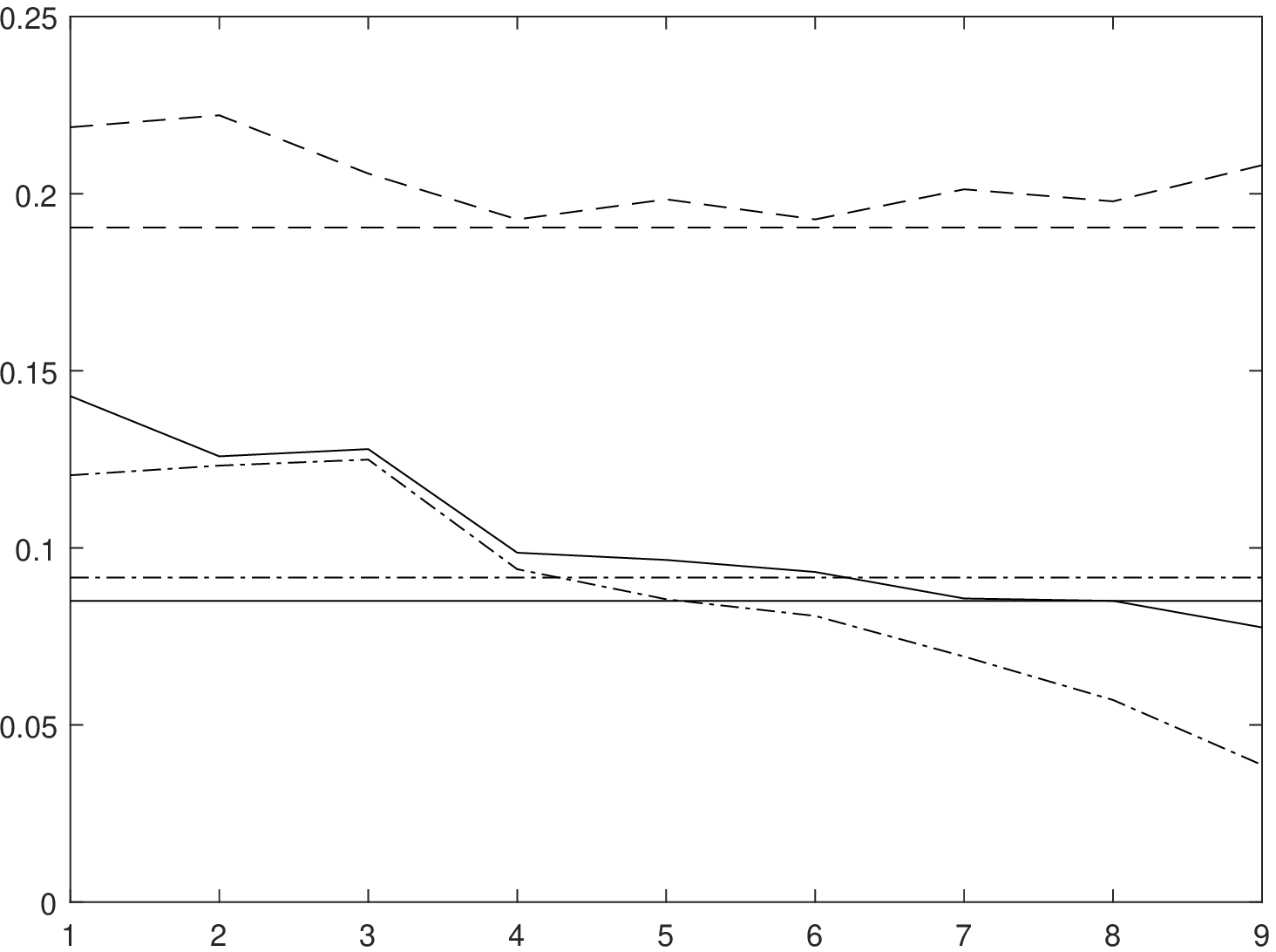}}
		\subfigure[Subject 4.]{\label{f4}\includegraphics[width=3.75cm]{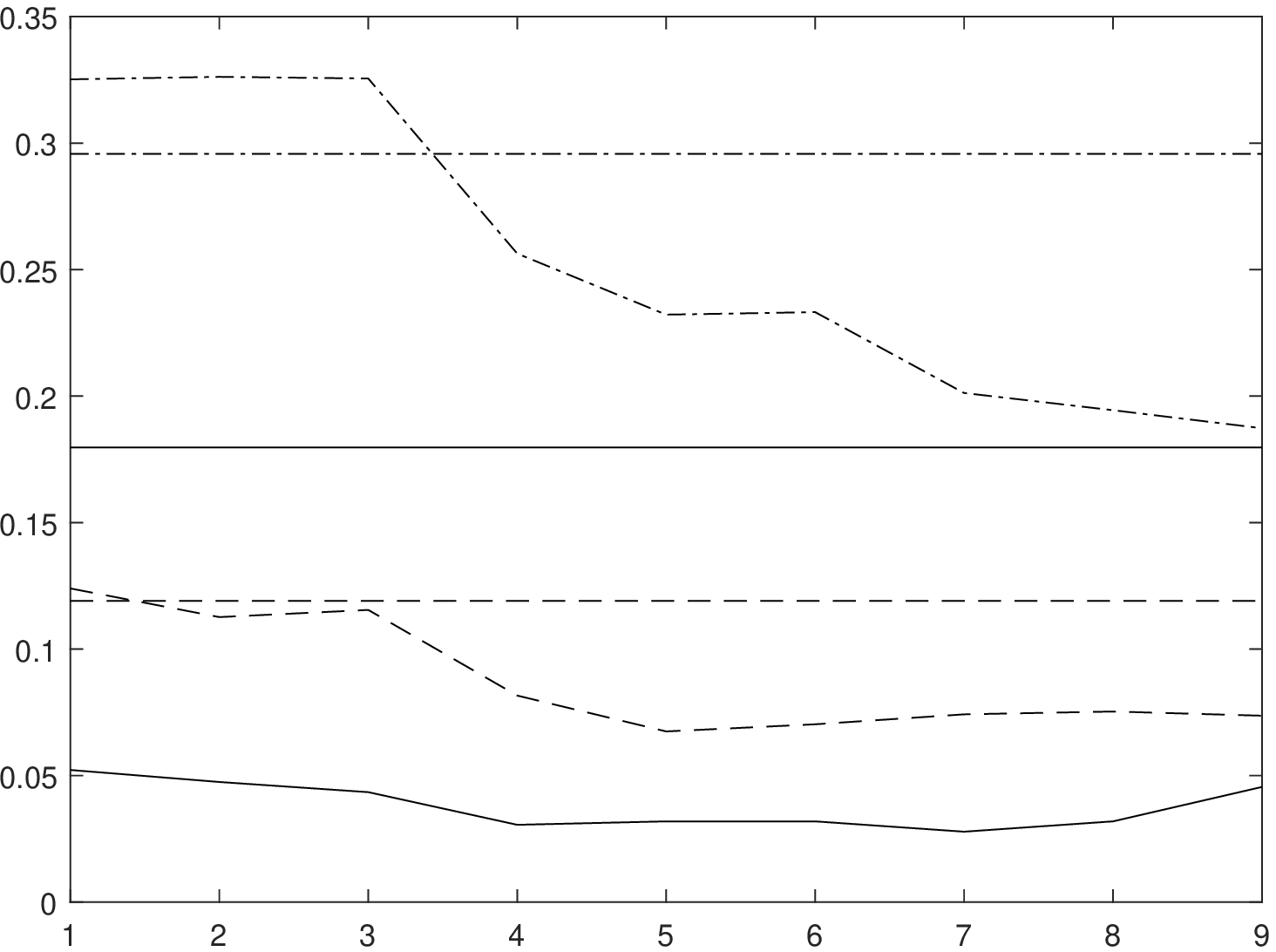}}
         \subfigure[Subject 5.]{\label{f5}\includegraphics[width=3.75cm]{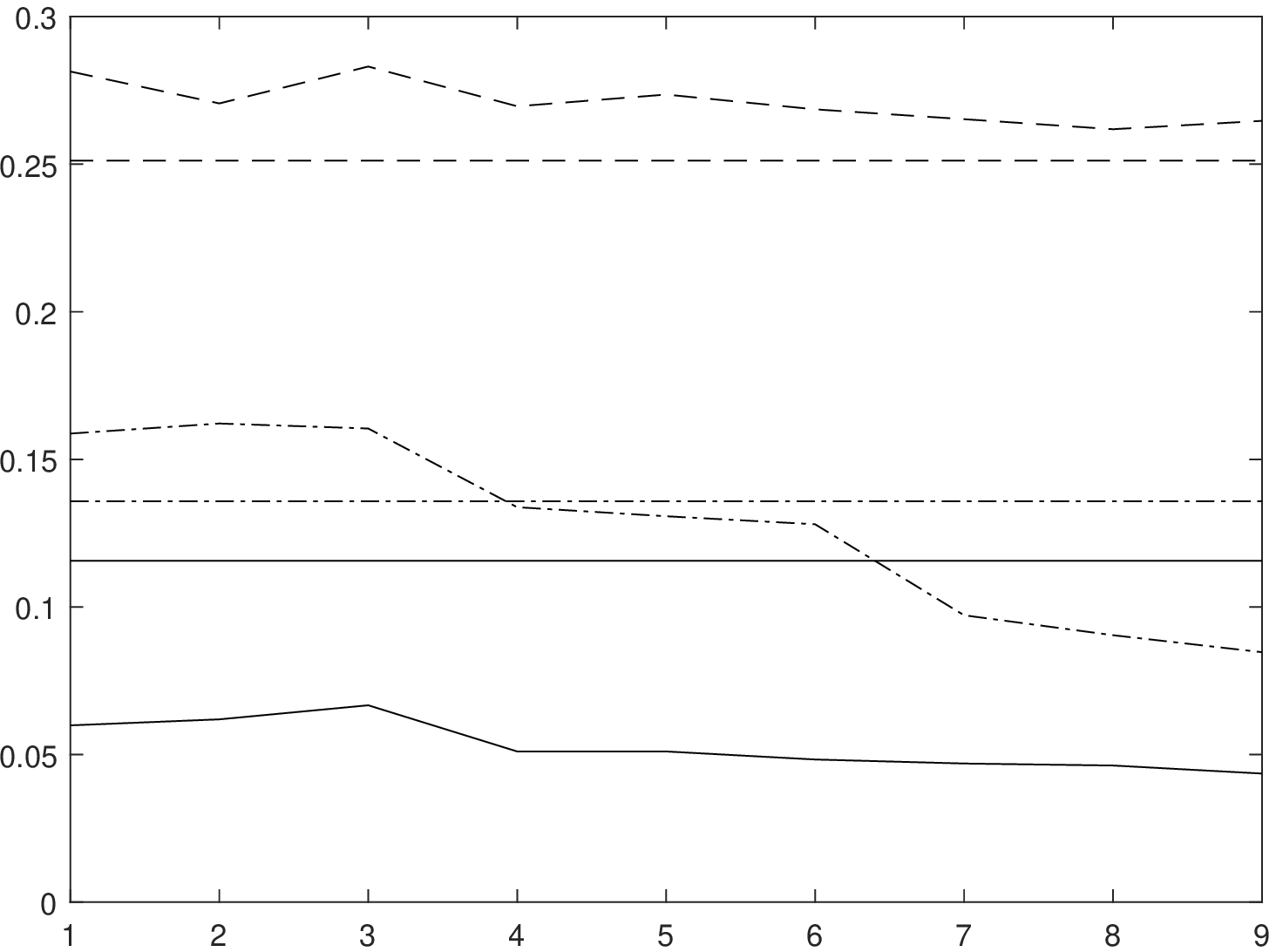}}
		\subfigure[Subject 6.]{\label{f6}\includegraphics[width=3.75cm]{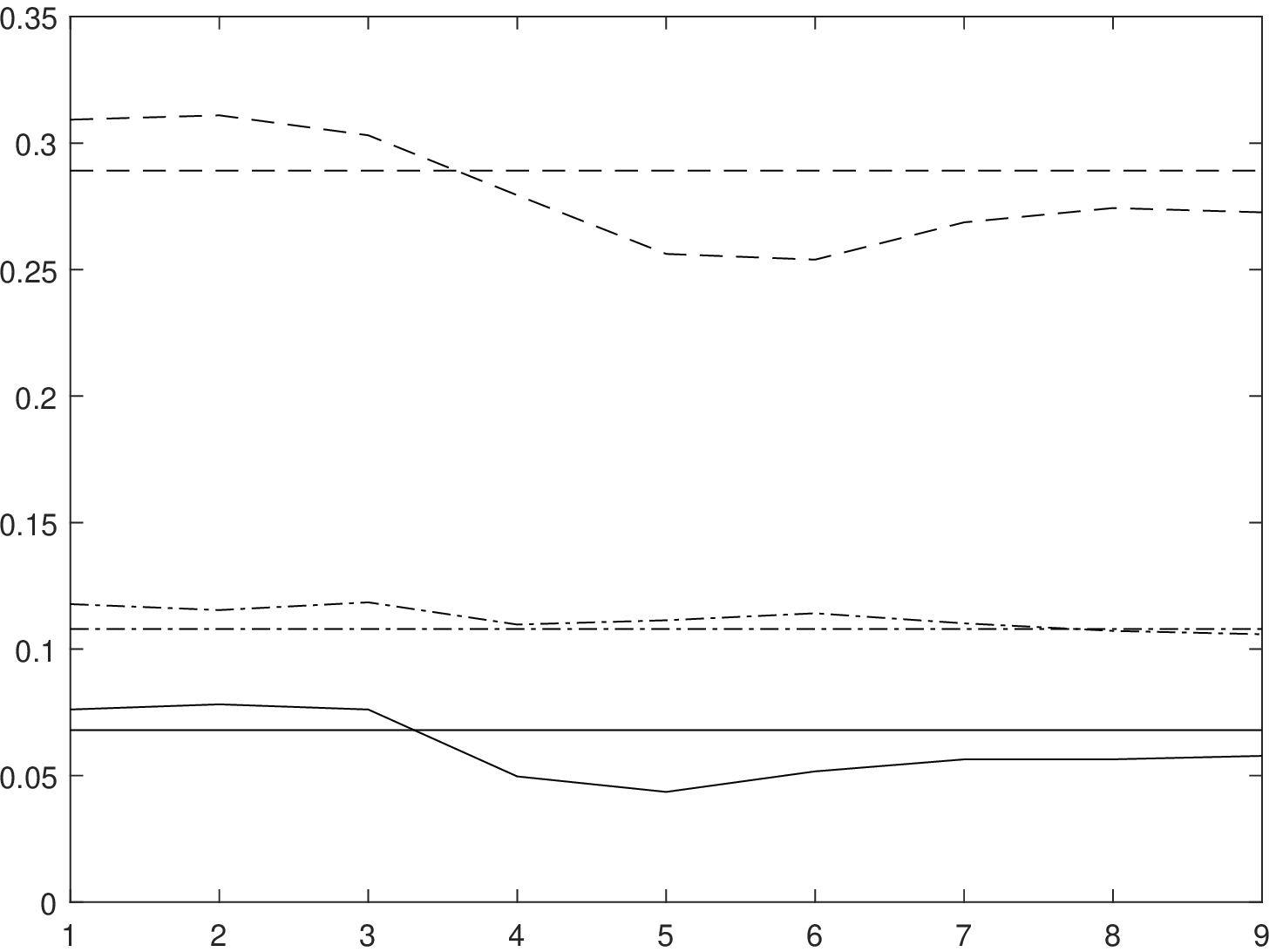}}
         \subfigure[Subject 7.]{\label{f7}\includegraphics[width=3.75cm]{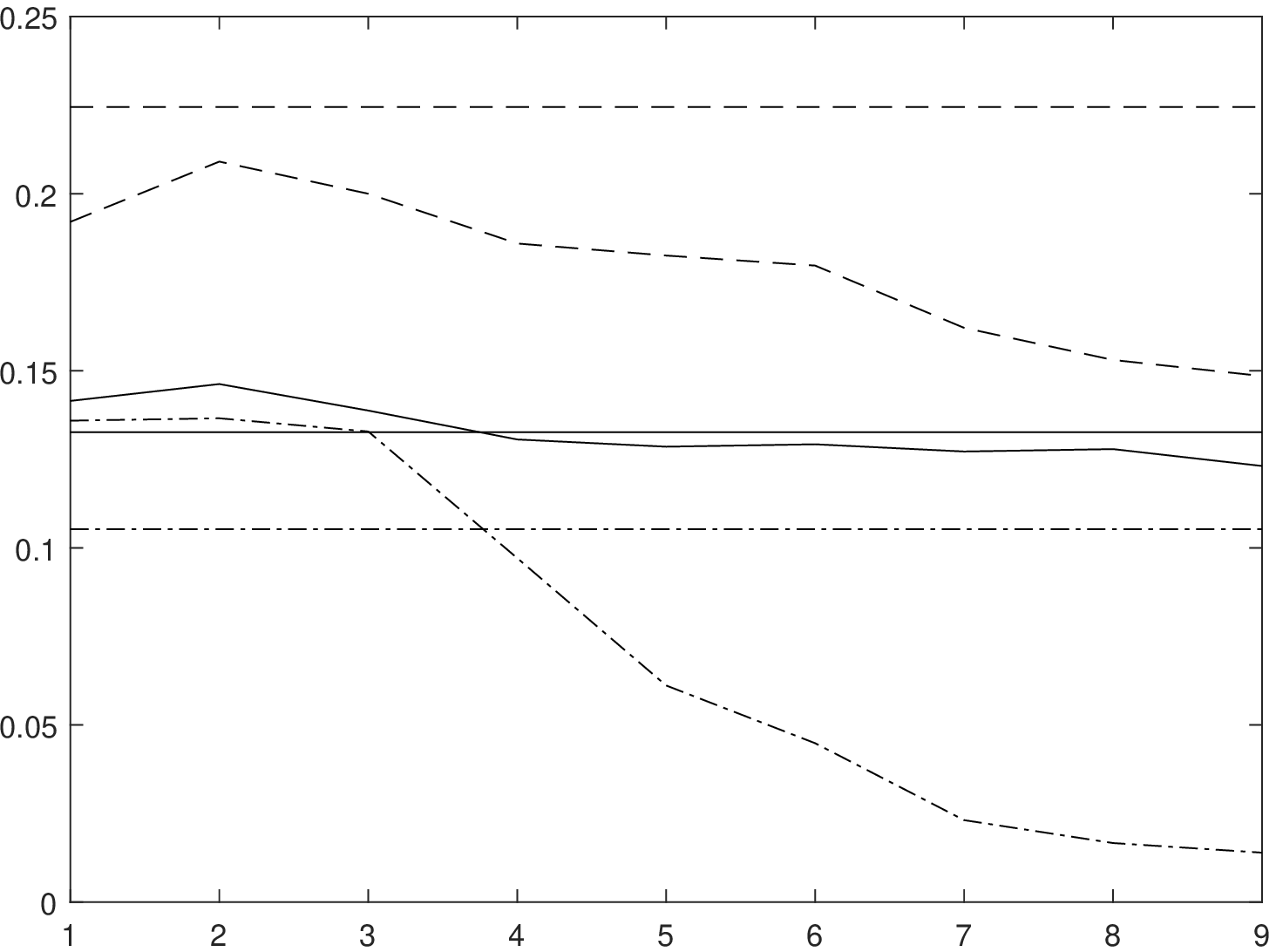}}
		\subfigure[Subject 8.]{\label{f8}\includegraphics[width=3.75cm]{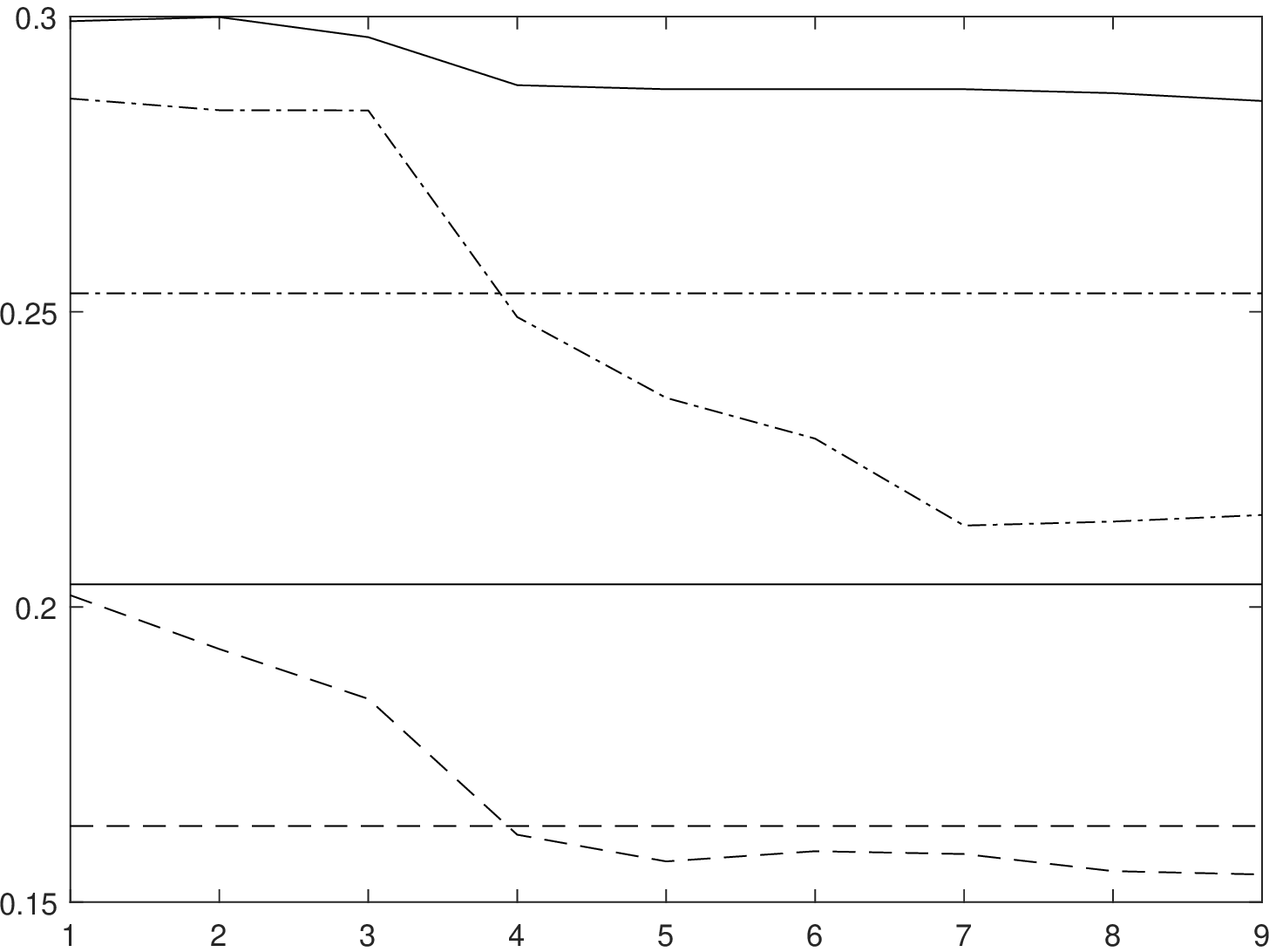}}
        \subfigure[Subject 9.]{\label{f9}\includegraphics[width=3.75cm]{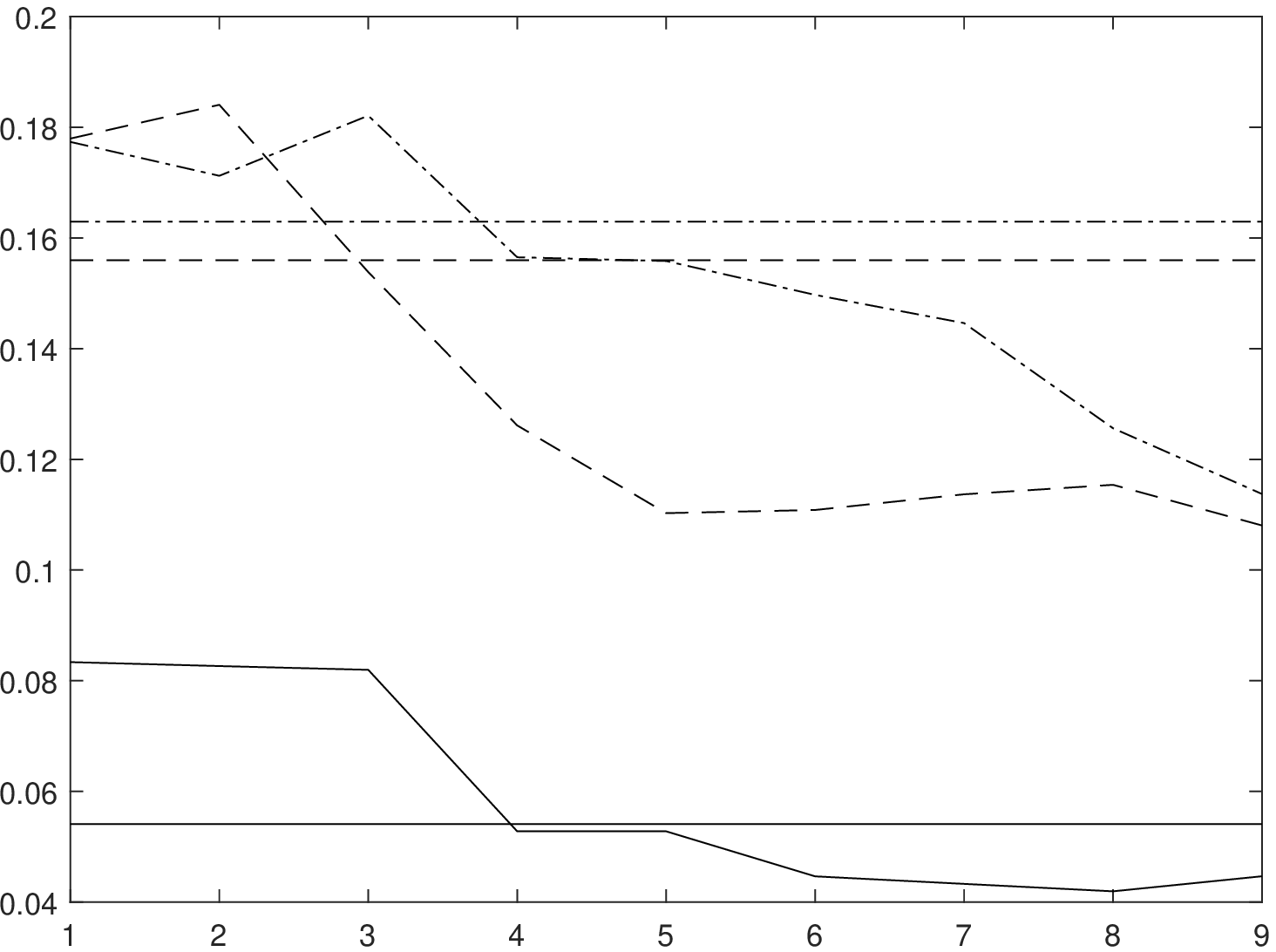}}
    \caption{Adding of new base models to Learn++ decreases the error rate. Error rate is shown in $y$-axis and $x$-axis shows the number of used base models. Thus, personalization improves the recognition rates compared to user-independent model (horizontal lines). Solid line shows the results of Learn++ with QDA, dashed line Learn++ with CART and dash-dotted line Learn++ with LDA.} 
\label{inc_learn}
\end{center}
\vspace{-2em}
\end{figure}

Experiments are made using leave-one-out method, in turn one person's data is used for personalization and testing and 8 persons' data for training the initial recognition model. The final model includes 9 base models, 3 user-independent and 6 personal. Three different base classifiers are compared: LDA, QDA and CART. Error rates were calculated for each 7 classes separately and mean of these was considered as the final error rate. Learn++ has random elements, and therefore, classification was performed 10 times for each subject. Average error rates ($y$-axis) from these are shown in Figure \ref{inc_learn}. As comparison, results using user-independent model were calculated. These were also trained using leave-one-out approach, in turn one person's data was left for testing and training was based on eight persons' data. Test person's data was divided to three parts, as explained in Figure \ref{process} and user-independent model was tested using only the last part which was also used to test Learn++ method. User-independent model was trained using all the training data unlike Learn++ which randomly samples data used for training. Therefore, the error rate of the first base model of Learn++ is not the same as the error rate of user-independent model.

According to the results shown in Figure \ref{inc_learn}, in most cases personalization reduces the average error rate: when new base models are added to Learn++, error rates decreases ($x$-axis shows the number of used base models). In fact, the improvement is significant: QDA and CART improves results in 7 cases out of 9 and LDA with all study subjects. With CART the average error rate drops from 18.0\% to 15.7\% (13.1\% improvement), with LDA from 14.1\% to 9.5\% (33.1\% improvement), and with QDA from 11.1\% to 9.1\% (17.9\% improvement), when compared to user-independent model. Therefore, while the average error rate using QDA is the smallest, the biggest benefit from personalization can be achieved when LDA is used as a base classifier. 


While the improvement is big with some subjects, it can be seen from Figure \ref{inc_learn} that if the initial, user-independent, model cannot classify some person's data reliably, incremental learning can not improve the error rate or even weakens it (e.g. see subject's 3 and 6 using CART). This because in some cases user-independent model cannot recognize instances of some classes at all. Therefore, when new base models  for personalization are trained based on labels from user-independent model, all class labels are not present in the training process. Thus, also personalized models cannot recognize these classes, and therefore, personalization cannot decrease the error rate in such cases. For this reason, it remains important to study how to train reliable user-independent models.

The results show that incremental learning can be used to personalize human activity recognition models and already a small personal data set had a huge positive impact to the recognition accuracy. In fact, the used data set contained only 3 minutes of data from each activity from each person, and most likely more comprehensive personal data set would lead to even better recognition rates. 
To test this and show the the full potential of the proposed method, it should be tested with more extensive data set containing labeled data from several days, and preferably, from several months. This data set could also be used to test how the presented method can handle changing situations.





\section{Conclusions}
\label{conclusion}

In this study, Learn++ was applied to human activity recognition data collected using smartphone sensors. Three base classifiers were compared: LDA, QDA and CART. Experiments showed that already a small personal training data set can improve the recognition accuracy significantly. Improvement using LDA as base classifier was 4.6 percentage units, using QDA 2.0 percentage units, and 2.3 percentage units using CART. However, it was noted that if the user-independent model used in the first phase of the recognition process is not accurate enough, personalization cannot improve recognition rates. Moreover, to show the full potential of the presented method, future work includes experimenting with more extensive data sets.

\begin{footnotesize}

\end{footnotesize}

\end{document}